\documentclass{article}

\usepackage{microtype}
\usepackage{graphicx}
\usepackage{subcaption}
\usepackage{booktabs}
\usepackage{tabularx}
\usepackage{placeins}
\usepackage[most]{tcolorbox}
\usepackage{hyperref}
\usepackage{fontawesome5}

\usepackage[accepted]{icml2026}

\makeatletter
\long\def\ICML@appearing{\textit{Proceedings of the AI for Law Workshop
  at the\/ $\mathit{43}^{rd}$ International Conference on Machine
  Learning}, Seoul, South Korea. PMLR 356, 2026. Copyright 2026 by
  the author(s).}
\makeatother

\usepackage{amsmath}
\usepackage{amssymb}
\usepackage{mathtools}
\usepackage{amsthm}

\usepackage[capitalize,noabbrev]{cleveref}

\theoremstyle{plain}

\theoremstyle{definition}

\theoremstyle{remark}

\usepackage[disable,textsize=tiny]{todonotes}

\icmltitlerunning{L-MARS: Legal Multi-Agent System with Citation-Faithfulness Audit}

\begin{document}

\twocolumn[
  \icmltitle{L-MARS: Legal Multi-Agent System with Agentic Search \\
    and Citation-Faithfulness Audit}

  \icmlsetsymbol{equal}{*}

  \begin{icmlauthorlist}
    \icmlauthor{Boqin Yuan}{ucsd}
    \icmlauthor{Ziqi Wang}{usc}
  \end{icmlauthorlist}

  \icmlaffiliation{ucsd}{University of California, San Diego, USA}
  \icmlaffiliation{usc}{University of Southern California, USA}

  \icmlcorrespondingauthor{Boqin Yuan}{b4yuan@ucsd.edu}

  \icmlkeywords{Legal AI, Citation Faithfulness, Retrieval-Augmented Generation, Agentic Search, AI4Law}

  \vskip 0.3in
]

\printAffiliationsAndNotice{}

\begin{abstract}
Large language models are increasingly deployed for legal question answering, where evaluations typically focus on multiple-choice accuracy. This measure overlooks a common failure: whether the citation source attached to an answer exists and supports the rule the system attributes to it. We present \textbf{L-MARS}, an open multi-agent legal QA system with agentic search and judge-driven evidence checks, and audit it claim by claim against its cited source. Each atomic claim is labelled with a six-class taxonomy and scored with strict-ALCE under cross-provider judging, where the answerer and verifier come from different model families. On a stratified 100-question Bar Exam audit, retrieval barely moves accuracy, yet the multi-turn judge loop lifts strict citation $F_1$ from $0.13$ (naive RAG) to $0.25$ and cuts the no-citation rate from $34\%$ to $13\%$. We further introduce \textbf{Faith-Search}, a post-draft step that re-verifies and repairs unreachable citations; it drops the unreachable rate below $1\%$ but does not improve $F_1$ over the multi-turn loop, so we report it as a targeted reachability intervention rather than a faithfulness breakthrough. A 50-question \textbf{LegalSearchQA} case study confirms the picture: retrieve-then-draft pipelines saturate near $0.75$ citation $F_1$, while a single-agent web-search baseline collapses to $0.22$ under external audit. We release code and prompts at \href{https://github.com/boqiny/L-MARS}{\faGithub\,\texttt{L-MARS}} and the LegalSearchQA dataset at \href{https://huggingface.co/datasets/boqiny/LegalSearchQA}{\raisebox{-0.2ex}{\includegraphics[height=1em]{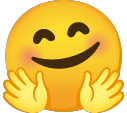}}\,\texttt{LegalSearchQA}}.
\end{abstract}

%%%%%%%%%%%%%%%%%%%%%%%%%%%%%%%%%%%%%%%%%%%%%%%%%%%%%%%%%%%%%%%%
\section{Introduction}
\label{sec:intro}

Large language models (LLMs) such as GPT \citep{openai2025gpt5systemcard}, Claude \citep{anthropic2025claudeopus4}, and Gemini \citep{deepmind2025gemini3} are increasingly applied to legal tasks including case law retrieval \citep{locke2022caselawretrievalproblems}, statutory interpretation, and automated legal assistance \citep{rincon2021legalclinic}. Direct use can produce hallucinations \citep{ji2023survey,xu2025hallucinationinevitableinnatelimitation}: the model states incorrect legal facts with high confidence. In legal contexts these errors carry real-world risk. Wrong citations and outdated statutes can undermine credibility and mislead decisions.

Domain-specific fine-tuning \citep{chalkidis2020legalbert} can help but requires frequent retraining as laws change. Retrieval-Augmented Generation (RAG) \citep{lewis2020rag} avoids retraining by fetching relevant documents at inference time, but its effectiveness depends on retrieval quality. Recent work has shown that on legal benchmarks centered on rule application, standard retrieval pipelines can have little effect on downstream QA accuracy. On Bar Exam QA, even the best traditional retriever improves GPT-4o-mini accuracy by only 0.5 percentage points over zero-shot \citep{Zheng_2025}.

We present \textbf{L-MARS}, a multi-agent legal QA system designed for answer generation with auditable evidence. L-MARS combines online search (Serper \citep{serperapi}, with Tavily \citep{tavilyapi} as a fallback backend), offline retrieval using BM25 over local documents, and case law retrieval through the CourtListener API \citep{courtlistenerapi}. Unlike naive retrieve-and-generate pipelines, L-MARS uses a multi-agent system that decomposes queries into structured search intents, executes targeted retrieval, filters evidence through a Judge Agent, and synthesizes grounded answers.

\begin{figure*}[!htbp]
  \centering
  \includegraphics[width=\linewidth]{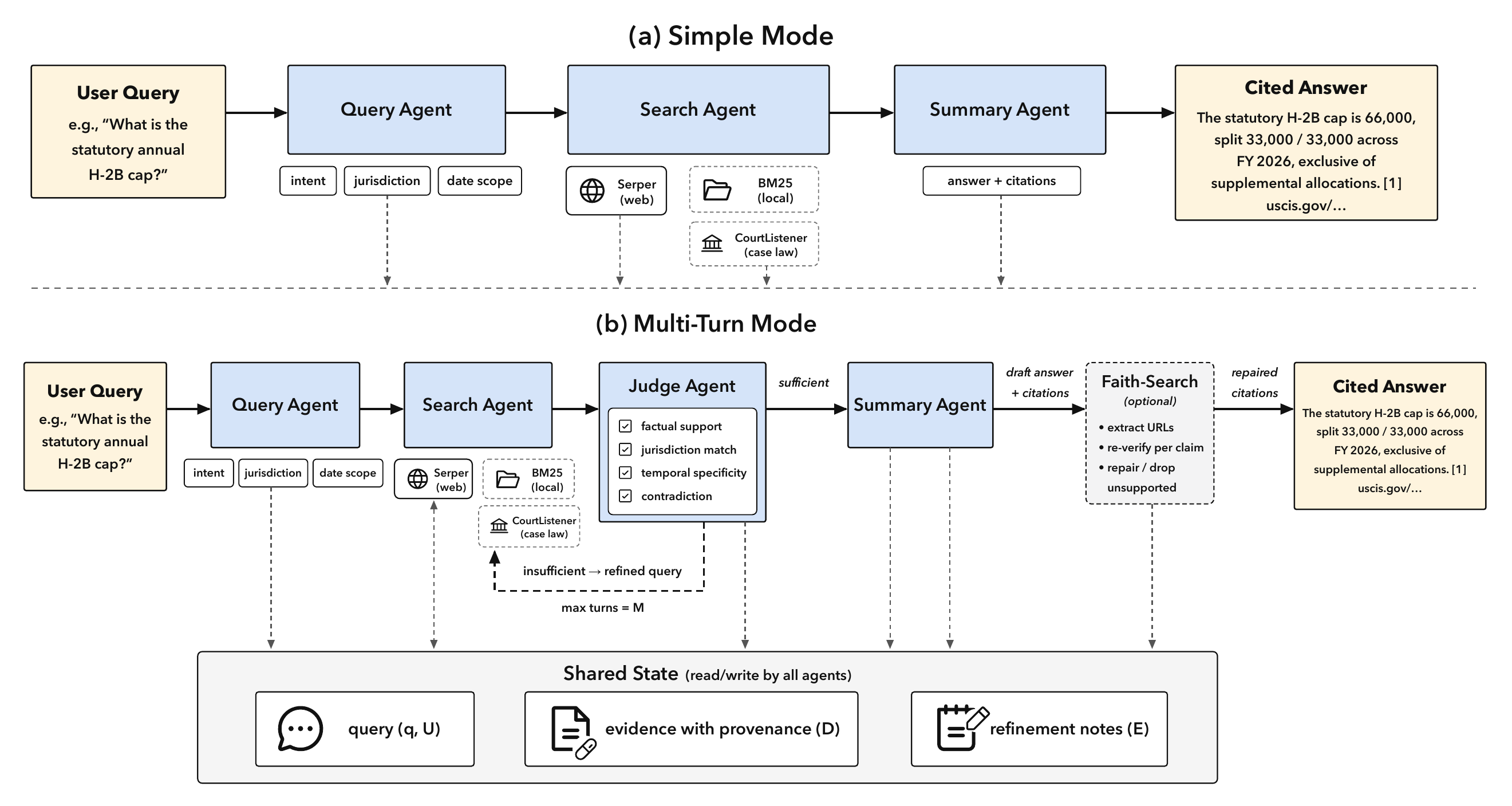}
  \caption{L-MARS system. Simple Mode runs Query, Search, and Summary agents once over web, local BM25, and CourtListener retrieval. Multi-Turn Mode adds a Judge Agent that checks evidence sufficiency and sends \texttt{insufficient} cases back for re-search, up to $M$ turns. The dashed Faith-Search block is the post-draft URL-repair extension evaluated as L-MARS-FS (\cref{sec:faithfulness}); panel (b) without it is L-MARS-MT.}
  \label{fig:workflow}
\end{figure*}

L-MARS supports a fast single-pass path and an iterative judge-driven path. The single-pass path decomposes the query, retrieves and filters evidence, and synthesizes a grounded answer. The iterative path repeats search, verification, and refinement until the evidence is sufficient or the retrieval budget is exhausted.

Answer accuracy alone is an incomplete measure of legal-QA quality. A system that picks the correct answer letter while citing fabricated authorities creates false trust in evidence that does not survive verification. Prior audits of closed commercial assistants \citep{magesh2025hallucinationfree,dahl2024legalfictions} report 17--33\% hallucinated citations despite retrieval, but to our knowledge no equivalent measurement exists for open multi-agent legal-QA systems, where each agent in the loop can introduce a fabricated reference.

The central observation that drives this paper is that legal QA should not treat citations as decorative explanations attached to an otherwise self-contained answer. We therefore evaluate systems at the level a legal user would inspect them: whether each answer claim is backed by a reachable source that actually supports it. This framing turns retrieval from a question of marginal accuracy gains into a question of evidentiary accountability, and explains why the systems we evaluate improve citation support far more than they improve answer accuracy. We audit L-MARS Multi-Turn against three baselines in \cref{sec:faithfulness}, surface a previously hidden failure mode in its cited URLs, and study a post-draft URL-repair extension as one targeted intervention.

We focus on three contributions, in the order an AI4Law reader is most likely to care about them. \emph{First}, we audit citation faithfulness for an open multi-agent legal-QA system, a setting that, to our knowledge, prior work has not measured directly. The audit uses a six-class claim taxonomy and strict-ALCE precision-recall adapted for unreachable citations. We use cross-provider judging across three model families: \texttt{gpt-5.4-mini} as the answerer, \texttt{gemini-2.5-flash} as the verifier, and \texttt{claude-haiku-4-5} as the third-family cross-evaluator. The multi-turn judge loop in L-MARS raises strict citation $F_1$ from $0.131$ (naive RAG) to $0.254$, a paired gain whose 95\% bootstrap CI excludes zero. \emph{Second}, we design \textbf{Faith-Search} as one targeted intervention with a null $F_1$ effect on this benchmark: a post-draft URL-repair step the verifier triggers when a claim's cite is flagged unreachable. Faith-Search drops the unreachable-citation rate from 2.4\% to 0.9\% and reaches $F_1 = 0.287$ on Bar Exam, but the matched-budget $F_1$ gain over the multi-turn loop is not statistically distinguishable from zero on $n = 100$. We therefore report Faith-Search as a reachability intervention, not a general faithfulness improvement; an ablation shows URL replacement is the necessary active component while the failure-type-aware router contributes no measurable benefit. \emph{Third}, we release the system that this audit is run on, \textbf{L-MARS}, a multi-agent legal-QA system that combines structured query decomposition, agentic web search, judge-driven evidence checks, and cited-answer synthesis, together with a 50-question \textbf{LegalSearchQA} case study for time-sensitive legal questions. The case study supplies external validity for time-sensitive legal search beyond the Bar Exam audit's reasoning-focused regime. All prompts, evaluation scripts, and raw outputs are released for reproducibility.

\section{Related Work}
\label{sec:related}

\paragraph{Agentic RAG and search systems.} Retrieval-augmented generation grounds model output in external evidence to reduce hallucination \citep{lewis2020rag}; interleaving reasoning with retrieval \citep{yao2023reactsynergizingreasoningacting,khattab2023} improves factual grounding further. Multi-agent frameworks \citep{wu2023autogen,langgraph2024} enable role specialization and iterative critique, and recent agentic-RAG variants combine planning, verification, and retrieval for long-horizon tasks \citep{singh2025agenticRAG,shinn2023reflexion}. On the search side, Search-o1 \citep{searcho1} and WebThinker \citep{li2025webthinkerempoweringlargereasoning} integrate agentic search into reasoning models; OpenAI's Deep Research \citep{openai2025deepresearch} exposes a production-grade web-scale research loop. These systems sit alongside reasoning models \citep{openai2024o1arxiv,openai2025o3systemcard,qwq-32b-preview,deepseek-r1} that scale deliberation at test time through chain-of-thought prompting \citep{wei2023chainofthoughtpromptingelicitsreasoning}, self-consistency \citep{wang2022selfconsistency}, and verifier-based calibration \citep{kadavath2022know}. L-MARS inherits the search-and-judge design pattern from Search-o1 and the agent-orchestration pattern from AutoGen and LangGraph, instantiated for the legal domain; our distinct contribution is the citation-faithfulness audit applied to this class of system, which prior work has not measured in the open agentic setting.

\paragraph{Legal AI and benchmarks.} Legal AI evaluation spans benchmarks for reasoning \citep{guha2023legalbench}, multi-task NLU \citep{chalkidis2022lexglue}, Chinese legal knowledge \citep{fei2023lawbench}, and pretraining data \citep{henderson2022pileoflaw}. More recent work targets retrieval and reasoning specifically: Bar Exam QA \citep{Zheng_2025} captures realistic legal RAG settings where standard retrieval fails, and LEXam \citep{fan2025lexambenchmarkinglegalreasoning} benchmarks multi-step legal reasoning across languages. \citet{katz2024gpt4bar} report that GPT-4 zero-shot passes the multistate bar exam, providing the multiple-choice methodology we adopt. \citet{zhang2025overruled} document that LLMs fail to track the temporal validity of legal authorities (for example, whether a precedent has been overruled), reinforcing the case for explicit retrieval on time-sensitive legal questions. On the systems side, ChatLaw \citep{cui2024chatlaw} explores multi-agent approaches for legal tasks, and \citet{steenhuis2025fetch} applies ensemble LLM classifiers to civil legal intake and referral, illustrating the deployment regime in which a retrieval-grounded answer assistant would sit. Many LegalBench \citep{guha2023legalbench} subtasks are now saturated (GPT-4o-mini reaches 100\% on \texttt{rule\_qa}), limiting their discriminative value for knowledge-intensive legal QA.

\paragraph{Legal hallucination and citation evaluation.} \citet{dahl2024legalfictions} document 58--88\% hallucination rates from vanilla LLMs on case-law tasks. \citet{magesh2025hallucinationfree} audit Lexis+ AI, Westlaw AI, and Practical Law's Ask AI assistant and find 17--33\% hallucinated citations \emph{despite} commercial retrieval, with failure modes concentrated in source identity rather than reasoning errors. \citet{bohnet2022attributedqa} formalize Attributed QA, and \citet{gao2023alce} introduce ALCE's strict citation precision and recall, which we adopt verbatim. \citet{wallat2024faithfulness} separate correctness from faithfulness in RAG attributions. Selective-classification frameworks \citep{geifman2017selective,gandelsman2025abstentionbench} provide standard abstention metrics. Our citation-faithfulness analysis (\cref{sec:faithfulness}) extends these tools to open multi-agent legal QA, a setting prior work has not directly audited.

\section{Methodology}
\label{sec:method}

\subsection{Problem Formulation}
Given a user query $q$ and optional clarifications $U$, the system maintains an evidence set $D$ and iterates between expanding $D$ (issuing new search queries) and synthesizing a cited answer $a$ with a reasoning chain $R$ that interleaves reasoning steps with retrieved evidence. A Judge Agent emits a sufficiency decision $s \in \{\textsc{Sufficient}, \textsc{Insufficient}\}$ and refinement notes $E$ that guide the next query state; the loop terminates when $s = \textsc{Sufficient}$ or after $M$ iterations.

\subsection{System and Agents}
L-MARS is implemented in \texttt{LangGraph} \citep{langgraph2024} with structured output validation and tool use. State transitions are deterministic and type-checked, so an answer trace is reproducible from the same query and a fixed seed. The system has four agents:

\paragraph{Query Agent.} Parses the user's question into a structured query containing intent, jurisdiction, date scope, and one or more search subqueries. Each subquery is annotated with a confidence and a triggering rationale.

\paragraph{Search Agent.} Executes tool calls to the retrieval backends and returns a list of normalized \texttt{SearchResult} objects (title, URL, snippet, optional full content, source type). The agent dispatches across web search (Serper), local BM25, and case-law retrieval (CourtListener) and applies a deduplication and provenance check before returning results to shared state.

\paragraph{Judge Agent.} Performs an explicit evidence-sufficiency checklist with $T = 0$: factual support, jurisdiction match, temporal specificity, and contradiction analysis. The agent returns \textsc{Sufficient} when the evidence answers each checklist item or \textsc{Insufficient} together with structured refinement notes used to seed the next iteration's query state.

\paragraph{Summary Agent.} Composes the final cited answer. Each atomic factual claim is paired with an inline citation that names the cited URL and snippet, so the answer is verifiable against retrieved evidence.

\subsection{Agentic Search}

\paragraph{Online search.} We adopt an agentic retrieval policy inspired by Search-o1 \citep{searcho1}: targeted web queries are interleaved with stepwise reasoning rather than issued in a single problem-oriented fetch. The Search Agent triggers retrieval when the Judge Agent flags a knowledge gap or the Query Agent predicts that external evidence will reduce uncertainty. Two modes are available: a fast \emph{basic search} that parses organic results into the normalized schema, and an \emph{enhanced search} that fetches page contents and PDFs and applies snippet-anchored extraction (token-level $F_1$ overlap, 2.5k-character window) to preserve local evidence while discarding boilerplate.

\paragraph{Local BM25.} We maintain a local retrieval index over user-provided documents. Markdown files are segmented into overlapping 500-character windows with 100-character stride and indexed with BM25 \citep{robertson2009bm25}. New files added to the input directory are picked up on the next query.

\paragraph{CourtListener.} The CourtListener Legal Search API provides access to U.S. judicial opinions, oral arguments, and dockets. When enabled, the Search Agent queries by party name, citation, docket number, or keyword and normalizes the API response into the same \texttt{SearchResult} schema as the other backends.

\subsection{Operating Modes}

\paragraph{Simple Mode.} A single forward pass with minimal latency: Query Agent $\to$ Search Agent $\to$ Summary Agent. No Judge Agent, no re-search loop. We use Simple Mode for the LegalSearchQA case study and the Bar Exam accuracy headline.

\paragraph{Multi-Turn Mode.} Adds a Judge Agent and an iterative search-judge-refine loop. The Query Agent may propose clarifying follow-ups (user responses, when available, are folded into the query state). Each iteration generates targeted search queries, runs enhanced search (top-3 with content), and invokes the Judge to decide sufficiency. If the Judge returns \textsc{Insufficient} the refinement notes are folded into the next query state and the loop repeats, up to a fixed maximum of $M$ turns. Multi-Turn Mode is the configuration audited for citation faithfulness in \cref{sec:faithfulness}.

\section{Experiments}
\label{sec:experiments}

\subsection{Tasks and Datasets}

\paragraph{LegalSearchQA (case study).} LegalSearchQA is a 50-question multiple-choice case study for time-sensitive legal questions, designed to test external validity of the systems beyond the Bar Exam audit's reasoning-focused regime rather than to serve as a benchmark. Questions span Tax, Corporate \& Financial Regulation, Labor \& Employment, Immigration, Technology \& Privacy, and Criminal, Drug \& State Law. Each item was authored and cross-referenced against the source most directly relevant to it during March 2026: predominantly primary government and judicial sources (e.g., IRS.gov, USCIS.gov, SEC.gov, the Federal Register, the Congressional Research Service, Supreme Court opinions; $\approx 70\%$ of items), with a minority of authoritative secondary sources (e.g., established law-firm publications, regulatory trackers; $\approx 30\%$). Correct answers require 2024--2026 legal or regulatory information.

\paragraph{Bar Exam QA.} We also evaluate on the 594-question Historical MBE subset of Bar Exam QA \citep{Zheng_2025}, covering Contracts, Torts, Constitutional Law, Evidence, Real Property, and Criminal Law. This benchmark emphasizes rule application over current information access: \citet{Zheng_2025} report that their best retriever improves GPT-4o-mini accuracy by only 0.5 points over zero-shot.

\subsection{Baselines and Conditions}

The LegalSearchQA case study and the Bar Exam citation-faithfulness audit use \texttt{gpt-5.4-mini}. Two auxiliary analyses use \texttt{gpt-4o-mini}: the full Bar Exam accuracy comparison, which follows the prior benchmark setting, and the LegalSearchQA per-domain diagnostic in the appendix. LegalSearchQA compares zero-shot, naive RAG, GPT-5.4-mini with native \texttt{web\_search}, L-MARS-MT, and L-MARS-FS. All agents use $T = 0$.

\subsection{Implementation}

For LegalSearchQA we evaluated all 50 questions under each condition. For Bar Exam QA we evaluated the full 594 questions under zero-shot and L-MARS, and we tested both Serper and Tavily retrieval backends. CourtListener was disabled because neither benchmark requires specific case citations.

%%%%%%%%%%%%%%%%%%%%%%%%%%%%%%%%%%%%%%%%%%%%%%%%%%%%%%%%%%%%%%%%
\section{Results}
\label{sec:results}

\subsection{Case Study: LegalSearchQA}
\label{sec:lsqa_case}

\begin{table*}[!htbp]
\centering
\footnotesize
\setlength{\tabcolsep}{4pt}
\caption{LegalSearchQA case study, $n = 50$ time-sensitive multiple-choice questions, \texttt{gpt-5.4-mini}\,$\times$\,\texttt{gemini-2.5-flash}. Reported as external validity for time-sensitive legal search, not as a benchmark-grade comparison. Brackets are 95\% bootstrap CIs (B = 10K, clustered by question, seed = 42).}
\label{tab:results_lsqa}
\begin{tabular}{lrlllrr}
\toprule
Condition & $n$ & Acc & $P_{\text{strict}}$ & $F_1$ strict & Unreach \% & No-cite \% \\
\midrule
Zero-Shot                     & 50 & 0.840\,[.74,.94] & 0.000\,[.00,.00] & 0.000\,[.00,.00] & 0.0  & 100.0 \\
Naive RAG                     & 50 & 0.900\,[.80,.98] & 0.749\,[.66,.83] & 0.747\,[.66,.83] & 0.0  & 0.5   \\
GPT-5.4-mini + web\_search    & 50 & 0.900\,[.80,.98] & 0.227\,[.12,.33] & 0.223\,[.12,.33] & 1.2  & 3.6   \\
L-MARS-MT                     & 50 & 0.920\,[.84,.98] & 0.736\,[.65,.82] & 0.722\,[.63,.80] & 2.7  & 3.8   \\
\textbf{L-MARS-FS}            & 50 & \textbf{0.920\,[.84,.98]} & \textbf{0.775\,[.69,.85]} & \textbf{0.767\,[.68,.85]} & 0.0  & 2.2   \\
\bottomrule
\end{tabular}
\end{table*}

\begin{figure*}[!htbp]
  \centering
  \includegraphics[width=0.92\linewidth]{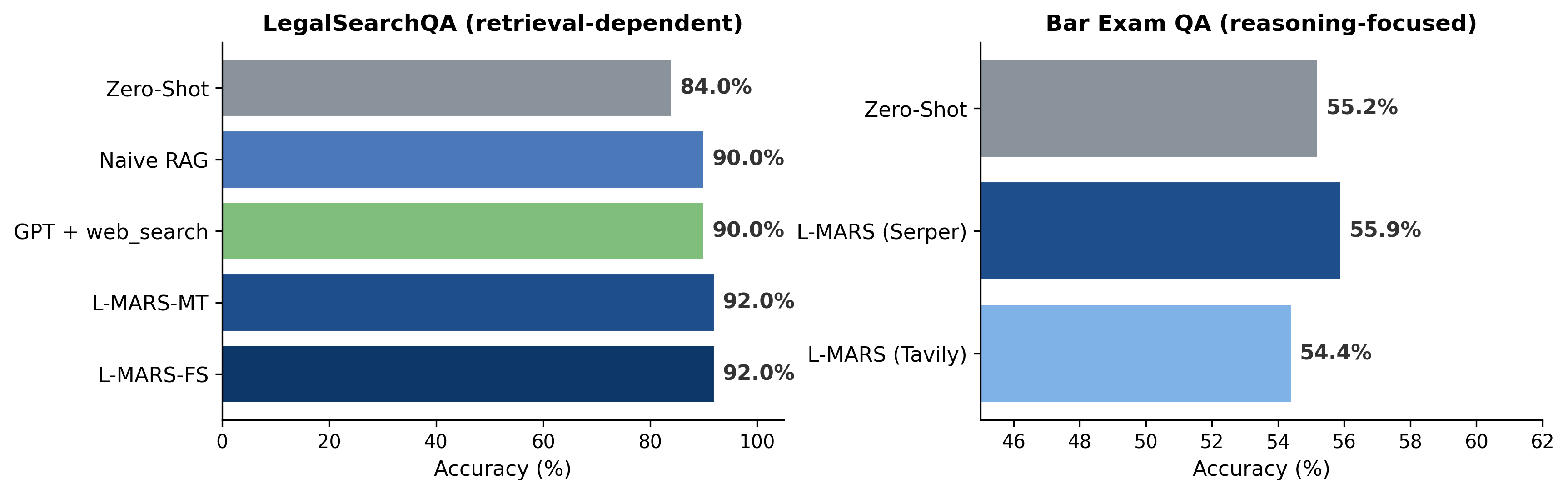}
  \caption{Two evaluation settings. \emph{Left:} on the retrieval-dependent LegalSearchQA case study every retrieve-then-draft pipeline saturates accuracy at 90--92\% while retrieval-blind Zero-Shot drops to 84\%. \emph{Right:} on reasoning-focused Bar Exam QA, retrieval provides negligible accuracy improvement over Zero-Shot regardless of the search backend. Accuracy is therefore an insensitive metric on Bar Exam; the multi-turn judge loop's effect on citation faithfulness, audited in \cref{sec:faithfulness} under this retrieval budget and verifier, is the signal of interest.}
  \label{fig:comparison}
\end{figure*}

\Cref{tab:results_lsqa} evaluates the LegalSearchQA case study with \texttt{gpt-5.4-mini} and the same citation verifier used in the Bar Exam audit. We read this case study as external validity for time-sensitive legal search, not as a benchmark-grade comparison; with $n = 50$ it sizes confidence intervals rather than power-runs hypothesis tests. Two patterns stand out. Accuracy saturates once retrieval is available: every retrieval-equipped system reaches 90--92\% and the CIs overlap. Citation faithfulness, in contrast, is sensitive to how the retrieved evidence is surfaced. Explicit retrieve-then-draft pipelines (Naive RAG, L-MARS-MT, L-MARS-FS) reach strict citation $F_1$ between $0.72$ and $0.77$, with overlapping CIs across the three; the single-agent native \texttt{web\_search} baseline drops to $F_1 = 0.223$ under external audit because most of its inline citations point to URLs whose externally-fetched content does not entail the claim. The case study is consistent with the Bar Exam picture in \cref{sec:audit_fs}: when retrieve-then-draft already keeps unreachable citations near zero, the post-draft URL-repair step in Faith-Search has little headroom, and the FS-over-MT gain is small. The older \texttt{gpt-4o-mini} per-domain run is reported in \cref{app:repro} as a domain-dependence diagnostic only.

\subsection{Bar Exam QA}
\label{sec:results_barexam}

\begin{table}[!h]
\centering
\small
\caption{Accuracy on Bar Exam QA (594 questions). All conditions use GPT-4o-mini.}
\label{tab:results_barexam}
\begin{tabular}{lcc}
\toprule
Condition & Acc.\ (\%) & Correct \\
\midrule
Zero-Shot       & 55.2 & 328/594 \\
L-MARS (Serper) & 55.9 & 332/594 \\
L-MARS (Tavily) & 54.4 & 323/594 \\
\midrule
\multicolumn{3}{l}{\textit{Prior work \citep{Zheng_2025}:}} \\
GPT-4o-mini      & 49.7 & -- \\
+ best retriever & 50.2 & -- \\
\bottomrule
\end{tabular}
\end{table}

\Cref{tab:results_barexam} reports the full 594-question Bar Exam accuracy comparison and \cref{fig:comparison} contrasts it side-by-side with the LegalSearchQA accuracy from \cref{tab:results_lsqa}. L-MARS provides a negligible 0.7-point gain with Serper and a slight 0.8-point degradation with Tavily, consistent with Bar Exam questions testing rule application more than current information access. On this benchmark, raw answer accuracy is therefore a poor discriminator; the signal we audit instead is citation faithfulness in \cref{sec:faithfulness}. Per-subject error patterns, LegalSearchQA misses, and retrieval-depth ablations are in \cref{app:repro}.

%%%%%%%%%%%%%%%%%%%%%%%%%%%%%%%%%%%%%%%%%%%%%%%%%%%%%%%%%%%%%%%%
\section{Citation-Faithfulness Analysis}
\label{sec:faithfulness}

\Cref{sec:results} shows that retrieval barely changes Bar Exam answer accuracy. The right question on this dataset is therefore not whether the system gets the letter right, but whether the evidence it cites exists and supports the claim it is used to justify. This section audits that question directly. We label every atomic claim in each answer against its cited source with a six-class taxonomy, report strict-ALCE precision and recall under cross-provider judging, and add a stress test against a blinded human pass on a stratified slice. \Cref{tab:headline,tab:cite_taxonomy} report the headline numbers; the rest of the section interprets them, runs ablations on the post-draft URL-repair step, and bounds the conclusions that survive the human stress test.

\begin{table*}[!htbp]
\centering
\small
\caption{Bar Exam citation audit, $n = 100$, \texttt{gpt-5.4-mini}\,$\times$\,\texttt{gemini-2.5-flash}. Brackets are 95\% bootstrap CIs (B = 10K, clustered by question, seed = 42).}
\label{tab:headline}
\begin{tabular}{lrlllrr}
\toprule
Condition & Acc & $P_\text{strict}$ & $R_\text{strict}$ & $F_1$ strict & Unreach \% & avg retr \\
\midrule
Zero-Shot   & 0.740\,[.65,.82] & 0.000\,[.00,.00] & 0.000\,[.00,.00] & 0.000\,[.00,.00] & 1.1 & 0    \\
Naive RAG   & 0.710\,[.62,.80] & 0.165\,[.11,.23] & 0.109\,[.07,.15] & 0.131\,[.09,.18] & 1.2 & 1    \\
\textbf{L-MARS-MT} & \textbf{0.760\,[.67,.84]} & \textbf{0.273\,[.22,.33]} & \textbf{0.237\,[.19,.29]} & \textbf{0.254\,[.21,.31]} & 2.4 & 2.66 \\
L-MARS-FS   & 0.690\,[.60,.78] & 0.370\,[.29,.45] & 0.234\,[.18,.30] & 0.287\,[.22,.35] & 0.9 & 1.72 \\
\bottomrule
\end{tabular}
\end{table*}

\begin{table*}[!htbp]
\centering
\small
\caption{Per-condition claim taxonomy (counts; percentages in parentheses). Bold cells highlight dominant failure modes for each condition.}
\label{tab:cite_taxonomy}
\begin{tabular}{lrrrrrr}
\toprule
Condition & claims & supported & partial & unsupported & unreachable & no\_citation \\
\midrule
Zero-Shot   & 646 & 0 (0.0\%) & 0 (0.0\%) & 0 (0.0\%) & 7 (1.1\%) & \textbf{639 (98.9\%)} \\
Naive RAG   & 589 & 64 (10.9\%) & 13 (2.2\%) & \textbf{303 (51.4\%)} & 7 (1.2\%) & 202 (34.3\%) \\
\textbf{L-MARS-MT} & 621 & \textbf{147 (23.7\%)} & 40 (6.4\%) & 336 (54.1\%) & 15 (2.4\%) & 82 (13.2\%) \\
L-MARS-FS   & 576 & 135 (23.4\%) & 15 (2.6\%) & 210 (36.5\%) & 5 (0.9\%) & 211 (36.6\%) \\
\bottomrule
\end{tabular}
\end{table*}

\subsection{Setup}
\label{sec:audit_setup}

We sample a stratified $n = 100$ subset of Bar Exam (seed = 42) and run four conditions on \texttt{gpt-5.4-mini}: zero-shot (no retrieval), naive RAG (single search query before the draft), L-MARS-MT (the multi-turn judge loop, $\leq 3$ retrievals), and L-MARS-FS, an extension of L-MARS-MT that adds a post-draft URL-repair step. All retrieval calls use Serper as the web search backend. For each atomic claim in the answer we run a citation verifier that issues one of six labels (\texttt{supported}, \texttt{partially\_supported}, \texttt{unsupported}, \texttt{citation\_unreachable}, \texttt{no\_citation}, \texttt{verifier\_error}) and we compute strict-ALCE precision $P_\text{strict} = S/(S+P+U+X)$, recall $R_\text{strict} = S/T$, and their harmonic mean, where $S$, $P$, $U$, $X$ are counts of \texttt{supported}, \texttt{partially\_supported}, \texttt{unsupported}, \texttt{citation\_unreachable} and $T$ is total claims. \texttt{verifier\_error} counts were zero across all four conditions, so the corresponding column is omitted from \cref{tab:cite_taxonomy} for compactness. Unreachable URLs are scored as failures (not abstentions), so a fabricated citation cannot game the metric.

\paragraph{Cross-provider judging.} The answering model is \texttt{gpt-5.4-mini} (OpenAI) and the citation verifier is \texttt{gemini-2.5-flash} (Google), different model family and different training data. The verifier reads only the cited source (canonical host plus path-prefix match, snippet fallback at token Jaccard $\geq 0.5$); it does not fall back to the rest of the retrieved pool. We additionally re-label a stratified $n = 57$ slice with \texttt{claude-haiku-4-5} as a third-family judge (\cref{sec:cross_eval}). A blinded human verification pass on an $n = 65$ slice (\cref{app:human_audit}) falls below our pre-registered $\kappa \geq 0.55$ release gate on the full slice; we use it as a stress test for the verifier rather than as validation of the labels.

\paragraph{What survives the human stress test.} Three observations bound the conclusions we draw. First, the stress test confirms that the LLM verifier and the human annotator agree most on the entailment buckets (supported, partially supported, unsupported), with $\kappa$ rising to $0.518$ on the $n = 35$ claims where neither side labeled \texttt{citation\_unreachable}. Second, the disagreement concentrates on URL reachability, a label that depends on fetch-time network state shared by neither pass; this is a property of point-in-time auditing rather than of the entailment judgment. Third, every comparison in this paper is therefore a relative comparison of systems under the same fixed verifier and audit window. We do not claim absolute citation-label correctness, and the $\Delta F_1$ values between L-MARS variants and the baselines are the quantities we ask the reader to trust.

\subsection{The Multi-Turn Loop Lifts Citation $F_1$}
\label{sec:audit_main}

\Cref{tab:headline} reports the main result. Naive single-pass RAG reaches strict citation $F_1 = 0.131$. Adding the multi-turn judge loop (L-MARS-MT) raises this to $0.254$, a paired $\Delta F_1 = +0.123$ with 95\% bootstrap CI $[+0.056, +0.191]$ that excludes zero. The judge loop helps in two coordinated ways visible in \cref{tab:cite_taxonomy}: the supported-claim count more than doubles (64 $\to$ 147 across $\sim 600$ claims), and the no-citation rate falls from 34\% to 13\% because the loop continues searching when the first query returns weak evidence.

The unreachable-citation rate is naturally low on this clean trace: $1.2\%$ for naive RAG and $2.4\%$ for L-MARS-MT. Both sit well below the $17$--$33\%$ band that \citet{magesh2025hallucinationfree} report for closed commercial assistants. An explicit retrieve-then-draft pipeline therefore already keeps fabricated-citation rates low, provided the search backend itself returns valid results.

\subsection{Faith-Search: A Targeted Reachability Intervention}
\label{sec:audit_fs}

L-MARS Faith-Search (L-MARS-FS) extends L-MARS-MT with two pieces: an atomic-claim verifier that scopes strictly to the cited source, and a URL-repair action that, when a claim's cite is flagged unreachable, replaces the dead cite with a newly-retrieved URL whose snippet still supports the claim. The pipeline reorganizes search so that the bulk of retrieval happens \emph{after} the first draft, in response to verifier flags. With a budget of $\leq 3$ total retrievals matched to L-MARS-MT, L-MARS-FS reaches $F_1 = 0.287$ (\cref{tab:headline}), a paired $\Delta F_1 = +0.033$ over L-MARS-MT with 95\% CI $[-0.048, +0.113]$ that overlaps zero.

We therefore describe Faith-Search as a targeted reachability intervention, not as a general faithfulness improvement. The unreachable rate drops from $2.4\%$ to $0.9\%$, which is the one outcome the URL-repair action is designed to produce. One side effect is visible in \cref{tab:cite_taxonomy}: L-MARS-FS shows a $36.6\%$ \texttt{no\_citation} rate, up from $13.2\%$ for L-MARS-MT. The mechanism is built into the repair branch: when URL repair cannot find a replacement source whose snippet still supports the claim, Faith-Search drops the citation rather than keep a broken one, which trades \texttt{citation\_unreachable} mass for \texttt{no\_citation} mass and cancels much of the F1 gain that the unreachable drop would otherwise have produced. The strict-$F_1$ gain over the multi-turn loop is not statistically distinguishable from zero on $n = 100$, because once L-MARS-MT already selects evidence whose host and snippet match the cited URL, post-draft URL replacement cannot create support that the underlying retrieval did not surface. Put plainly: URL repair fixes broken citations; it does not invent supporting evidence where none was retrieved. This motivates testing the intervention in regimes where the base system has many unreachable citations to begin with (commercial legal assistants, single-pass agents with light retrieval, the GPT-5.4-mini native \texttt{web\_search} baseline on LegalSearchQA), and we report its present-data behavior as an honest null result on the F1-over-MT comparison.

\subsection{Ablations: What Inside Faith-Search Is Doing the Work?}
\label{sec:audit_ablations}

Two ablations isolate which piece of Faith-Search matters, holding the retrieval budget fixed at $\leq 3$ calls.

\paragraph{A1a -- remove the failure-type-aware router.} Faith-Search picks specialized re-query strategies based on the verifier's failure type (authoritative-domain bias for fabricated URLs, date constraints for stale ones, gap-targeted reformulation for unsupported ones). A1a replaces this with a single generic claim-text re-query. Result: $F_1 = 0.286$ versus the full Faith-Search at $0.287$; the unreachable rate is even lower ($0.5\%$). The router contributes nothing measurable; the simple re-query suffices.

\paragraph{A1b -- disable URL replacement inside Faith-Search.} The verifier still flags unreachable citations, but the system does not substitute a working URL: flagged claims are simply marked as failed. Result: $F_1$ collapses to $0.170$ and the unreachable rate climbs back to $2.3\%$. Without replacement, the verifier only counts failures and never fixes them.

Within Faith-Search and at this retrieval budget, URL replacement is the necessary component: removing it (A1b) drops $F_1$ by $0.117$ and reverts the unreachable rate from $0.9\%$ to $2.3\%$. The failure-type-aware router is dispensable: A1a, a single generic re-query, is statistically indistinguishable from the full Faith-Search at $F_1 = 0.286$ versus $0.287$. Faith-Search itself does not improve significantly over L-MARS-MT on this benchmark ($\Delta F_1 = +0.033$ with CI overlapping zero), so the ablations describe the internal mechanism of Faith-Search, and the multi-turn baseline remains the stronger choice at this retrieval budget.

\subsection{Cross-Evaluator Robustness}
\label{sec:cross_eval}

Because the citation verifier is itself an LLM, we re-label a stratified $n = 57$ Case-3 slice (claims whose cite was located in the retrieved set, so the judge issued an NLI call) with \texttt{claude-haiku-4-5} as a third-family judge. 3-class Cohen's $\kappa$ between Gemini and Claude-Haiku is $\mathbf{0.683}$; strict-binary $\kappa$ is $\mathbf{0.661}$. Both clear the 0.55 release-gate threshold. Most disagreements stay within adjacent labels (supported $\leftrightarrow$ partial, partial $\leftrightarrow$ unsupported) and few cross the taxonomy more than one step. Two LLM judges from different families assigning consistent entailment labels is consistent with the labels carrying real signal, although it stops short of validation against non-LLM ground truth. A blinded human verification pass on an $n = 65$ slice falls below the pre-registered gate and is reported in \cref{app:human_audit} as a stress test.

\subsection{What This Audit Does Not Fix}
\label{sec:not_fixed}

The audit measures and partially addresses one failure mode: source-identity errors visible as unreachable citations. It does not move the larger \texttt{no\_citation} (37\% for L-MARS-FS) and \texttt{unsupported} (36\%) buckets. Pushing absolute $F_1$ above $0.5$ on Bar Exam will require separate work on cite-density-rewarding prompts and on stronger entailment-time retrieval. We discuss these and other limits in \cref{sec:conclusion}.

%%%%%%%%%%%%%%%%%%%%%%%%%%%%%%%%%%%%%%%%%%%%%%%%%%%%%%%%%%%%%%%%
\section{Conclusion}
\label{sec:conclusion}

We presented a citation-faithfulness audit of an open multi-agent legal-QA system, the audited system itself (L-MARS), and a 50-question LegalSearchQA case study for time-sensitive legal questions. The Bar Exam audit shows that the multi-turn judge loop in L-MARS raises strict citation $F_1$ from $0.131$ (naive RAG) to $0.254$, a gain whose 95\% bootstrap CI excludes zero, mainly by recovering supported citations and cutting the no-citation rate from 34\% to 13\%. A post-draft URL-repair extension, Faith-Search, addresses a narrower failure mode: it drops unreachable citations from 2.4\% to 0.9\% and reaches $F_1 = 0.287$, but its matched-budget $F_1$ lift over the multi-turn loop is not statistically distinguishable from zero on $n = 100$. We therefore report Faith-Search as a targeted reachability intervention rather than as a general faithfulness improvement; ablations show that URL replacement is the necessary component while the failure-type-aware router contributes no measurable benefit. On the LegalSearchQA case study, every retrieve-then-draft pipeline saturates accuracy and citation $F_1$, while a single-agent native web-search baseline collapses on externally auditable citations.

\paragraph{Implications for legal AI.} Accuracy alone is a misleading metric on Bar Exam: every retrieval setting we tested produces roughly the same letter-grade answer, and a system judged on accuracy alone looks fine even when its evidence is broken. At the claim level, on the audited traces and under the verifier we use, the multi-agent search loop roughly doubles how often the cited source actually supports the claim, although the absolute level of citation support remains modest. Three implications follow. First, legal QA assistants warrant evaluation on supportability of evidence as well as on whether the final letter is correct, because an accuracy-only evaluation hides citation breakage. Second, in this setting multi-agent retrieval helps supportability more than it helps raw accuracy, so the value of agentic search in law on these tasks lies primarily in producing answers whose evidence a legal user can verify line by line. Third, the audit itself is fragile: citation reachability drifts across fetch environments and time, so any production system that depends on URL-level verification needs an explicit policy for handling that drift. We demonstrate improved citation support in this setup; we do not claim a general solution to legal faithfulness.

\paragraph{Limitations.} LegalSearchQA is small (50 items) and time-sensitive. The Bar Exam audit uses a stratified $n = 100$ subset and a single backbone family (\texttt{gpt-5.4-mini}); the citation verifier is itself an LLM, so the cross-provider and human checks should be read as triangulation evidence and stop short of non-LLM ground truth. The human stress test, in particular, prevents us from claiming absolute correctness of any individual citation label, but it does not undercut the relative comparisons across systems under the same fixed verifier and audit window, which are the quantities the paper reports. The audit measures one failure mode (citation source identity); the larger \texttt{unsupported} (36--54\%) and \texttt{no\_citation} (13--37\%) buckets are not addressed by Faith-Search and dominate the residual error mass. Pushing absolute $F_1$ above 0.5 will require separate work on cite-density-rewarding prompts and stronger entailment-time retrieval.

%%%%%%%%%%%%%%%%%%%%%%%%%%%%%%%%%%%%%%%%%%%%%%%%%%%%%%%%%%%%%%%%

%%%%%%%%%%%%%%%%%%%%%%%%%%%%%%%%%%%%%%%%%%%%%%%%%%%%%%%%%%%%%%%%
\appendix
\onecolumn
\raggedbottom

\section{Agent Instructions}
\label{app:instructions}

\begin{tcolorbox}[
    breakable, enhanced jigsaw,
    colframe=gray, colback=gray!5!white, coltext=black,
    fonttitle=\bfseries, title=Query Agent,
    boxrule=1pt, arc=2mm, width=\linewidth,
    left=7pt, right=7pt, top=2pt, bottom=2pt]
\fontsize{8pt}{9.5pt}\selectfont
\begin{verbatim}
The Query Agent is a stateless pass-through that wraps the user
query into a structured QueryGeneration dataclass:

    @dataclass
    class QueryGeneration:
        query: str
        query_type: str = "web_search"
        priority: str = "high"

    class QueryAgent:
        def build_query(self, user_query: str) -> QueryGeneration:
            return QueryGeneration(query=user_query.strip())
\end{verbatim}
\end{tcolorbox}

\begin{tcolorbox}[
    breakable, enhanced jigsaw,
    colframe=gray, colback=gray!5!white, coltext=black,
    fonttitle=\bfseries, title=Summary Agent,
    boxrule=1pt, arc=2mm, width=\linewidth,
    left=7pt, right=7pt, top=5pt, bottom=5pt]
\fontsize{8pt}{9.5pt}\selectfont
\begin{verbatim}
You are a bar exam expert. A user will provide retrieved legal
sources and a multiple-choice question. Use the retrieved sources
together with your legal knowledge to select the single best
answer.

Rules:
1. Your response MUST start with exactly one letter on the first
   line: A, B, C, or D.
2. Follow with a concise 1-2 sentence rationale citing the
   controlling legal rule.
3. Bar exam questions test specific exceptions and nuances; do
   not just apply the most common general rule if the specific
   facts call for an exception.

Response format:
[Letter]
[Brief rationale]

The user message is structured as:
  Retrieved Evidence:
  <concatenated search results with source, title, content>
  Question:
  <full MCQ with answer choices A-D>
\end{verbatim}
\end{tcolorbox}

\begin{tcolorbox}[
    breakable, enhanced jigsaw,
    colframe=gray, colback=gray!5!white, coltext=black,
    fonttitle=\bfseries, title=Judge Agent,
    boxrule=1pt, arc=2mm, width=\linewidth,
    left=7pt, right=7pt, top=5pt, bottom=5pt]
\fontsize{8pt}{9.5pt}\selectfont
\begin{verbatim}
You are a legal research quality judge for US bar exam questions.

You will receive a bar exam multiple-choice question and a list of
retrieved search result snippets. Your job is to decide whether
the retrieved evidence is sufficient to reliably answer the
question.

"Sufficient" means: at least one result contains specific legal
rules, statutes, case holdings, or doctrinal exceptions that
directly address the core legal issue tested by the question;
not just a restatement of the facts.

If the evidence is NOT sufficient, provide a better search query
that targets primary legal sources (specific rule names, statute
citations, or doctrine names) rather than generic question text.

Respond with valid JSON only, no markdown fences:
{"sufficient": true_or_false,
 "reason": "one sentence",
 "next_query": "refined query (omit if sufficient)"}

Pre-filtering: before LLM evaluation, results matching any of
these patterns are deterministically removed at zero cost:
HTTP error, 403/404, Connection error, Request timed out,
JavaScript is disabled, No search results found.
\end{verbatim}
\end{tcolorbox}

\begin{tcolorbox}[
    breakable, enhanced jigsaw,
    colframe=gray, colback=gray!5!white, coltext=black,
    fonttitle=\bfseries, title=VerifyAgent (citation-faithfulness),
    boxrule=1pt, arc=2mm, width=\linewidth,
    left=7pt, right=7pt, top=5pt, bottom=5pt]
\fontsize{8pt}{9.5pt}\selectfont
\begin{verbatim}
DECOMPOSE: You decompose a legal answer into atomic factual
claims. Skip rhetorical sentences, conclusions, and pure
restatements of the question. Each atomic claim is a single,
self-contained factual assertion that can be independently
checked against a source.

ENTAIL (per claim): You are an NLI-style judge for a single
legal claim against ONLY its cited source. Return one of:
supported            -- the cited source NLI-entails the claim
partially_supported  -- a subset of predicates is entailed
unsupported          -- the cite is reachable but does not entail
citation_unreachable -- the cite is absent from the retrieved
                        pool, fabricated, or dead at external fetch
no_citation          -- the claim is made without a cite
verifier_error       -- judge failed to produce a label

Strict scoping: match the cited source by canonical-host plus
path-prefix, with a snippet-Jaccard >= 0.5 fallback. Do not
silently rescue a fabricated URL with adjacent retrieved evidence.
\end{verbatim}
\end{tcolorbox}

\section{Case Study: Post-Training Knowledge Gap}
\label{app:case-study}

Two representative items from LegalSearchQA where zero-shot and chain-of-thought both fail from stale parametric knowledge, while L-MARS retrieves the current information.

\begin{tcolorbox}[
    breakable, enhanced jigsaw,
    colframe=gray, colback=gray!5!white, coltext=black,
    fonttitle=\bfseries, title=Q1 (lsqa\_001): Executive Order 14110 status,
    boxrule=1pt, arc=2mm, width=\linewidth,
    left=7pt, right=7pt, top=5pt, bottom=5pt]
\fontsize{8pt}{9.5pt}\selectfont
\begin{verbatim}
Question: What is the current status of Executive Order 14110,
the Biden administration's executive order on AI safety signed
in October 2023?

A. It remains in full effect and is being actively enforced
B. It was revoked by President Trump on January 20, 2025
C. It was partially amended but key provisions remain
D. It expired automatically after 12 months

Correct Answer: B

Zero-Shot selected A. The model's parametric knowledge has EO
14110 as an active, enforceable order; the January 2025
revocation post-dates training data.

Chain-of-Thought selected A. The model reasons step-by-step from
stale knowledge: it recalls that EO 14110 was signed in October
2023, notes that executive orders do not expire automatically,
and concludes it must still be in effect. The reasoning is
internally consistent but factually wrong.

L-MARS selected B. Retrieved sources included whitehouse.gov and
Federal Register entries confirming revocation via EO 14148 on
January 20, 2025.
\end{verbatim}
\end{tcolorbox}

\begin{tcolorbox}[
    breakable, enhanced jigsaw,
    colframe=gray, colback=gray!5!white, coltext=black,
    fonttitle=\bfseries, title=Q2 (lsqa\_002): 2025 standard deduction,
    boxrule=1pt, arc=2mm, width=\linewidth,
    left=7pt, right=7pt, top=5pt, bottom=5pt]
\fontsize{8pt}{9.5pt}\selectfont
\begin{verbatim}
Question: For tax year 2025, what is the standard deduction for
a single filer under age 65, after the One Big Beautiful Bill
Act amendments?

A. $14,600    B. $15,000    C. $15,350    D. $15,750

Correct Answer: D

Zero-Shot selected C ($15,350): the model recalls an intermediate
figure from pre-OBBBA IRS projections but lacks knowledge of the
legislative amendment.

Chain-of-Thought selected C ($15,350): step-by-step reasoning
reinforces the wrong figure -- the model recalls the 2024 standard
deduction ($14,600), applies an estimated inflation adjustment,
and arrives at $15,350. The reasoning is plausible but anchored
in stale data.

L-MARS selected D ($15,750): retrieved current IRS sources
confirming the One Big Beautiful Bill Act raised the 2025
standard deduction to $15,750 for single filers.
\end{verbatim}
\end{tcolorbox}

Both items illustrate the same failure mode: the model's parametric knowledge is confidently wrong about post-training developments. Chain-of-thought \emph{amplifies} the failure: by encouraging step-by-step reasoning from stale premises, CoT produces internally coherent but factually incorrect justifications. L-MARS resolves both cases by retrieving current authoritative sources (whitehouse.gov, IRS.gov) and grounding the answer in retrieved evidence rather than parametric memory.

\section{Human Verification Audit}
\label{app:human_audit}

We collected a blinded human audit on a stratified $n = 65$ subset (5 claims per (condition $\times$ label) cell, seed $= 42$, drawn across all four conditions). The annotator saw only the claim, the cited URL, and the cited snippet, with no indication of which system produced the claim or which label the Gemini judge had assigned. A Claude Opus 4.7 agent with a web-fetch utility opened each cited URL and surfaced potentially relevant passages. The human annotator then read the surfaced text alongside the snippet, applied the four-label rubric released with the code, and entered the final label.

Claude's provisional labels (\texttt{slice\_annotated\_claude.csv}) and the final human labels (\texttt{slice\_annotated\_human\_final.csv}) are both released so the override trail is auditable; all $\kappa$ values reported in the paper are computed against the human-final CSV. The strict-binary $\kappa$ between the human-final labels and the Gemini judge is $\mathbf{0.458}$ on the full $n = 65$ slice, below our pre-registered 0.55 release gate; the 3-class $\kappa$ is $0.219$. Restricted to the $n = 35$ claims where neither side labeled \texttt{citation\_unreachable}, both $\kappa$ values rise to $\mathbf{0.518}$, within the moderate-agreement band.

The disagreement concentrates on \texttt{citation\_unreachable}: of 20 URLs the Gemini judge reached at audit time, 15 were unreachable when the human refetched them; of 15 URLs the human reached, 10 returned content the Gemini judge had earlier labeled unreachable. Three plausible causes contribute to this divergence: the two passes ran in different fetch environments (different user-agents, different proxy paths, no shared cache), the verifier and the human applied slightly different operational tests for \emph{absent from retrieved pool} versus \emph{dead at external fetch}, and URL reachability itself drifts between inference time and audit time. The result shows that anchoring an audit trail to a single point-in-time reachability check is brittle by construction. Because the pass uses one annotator and Claude-assisted source retrieval, we use it as a stress test for the verifier.

\section{Supplementary Analyses}
\label{app:repro}

\paragraph{Error analysis (LegalSearchQA).} L-MARS missed two questions. Q27 asked about H-1B weighted-selection details, a procedural rule about wage-level assignment; the retrieved results did not surface the specific Federal Register provision. Q44 asked for the exact effective date of T+1 settlement; search results discussed the policy broadly but missed the precise date. Both errors share a pattern: the correct answer requires a specific fact buried in a longer document, and snippet-level retrieval missed the critical sentence. The chain-of-thought baseline degrades to 30.0\% on this set because the model recalls outdated dollar amounts from prior tax years and builds multi-step justifications around them; this is the canonical confident-confabulation failure mode in any domain where facts change faster than model training cycles.

\paragraph{Per-domain breakdown (LegalSearchQA, GPT-4o-mini reference).} On the earlier GPT-4o-mini reference run, the largest retrieval-driven gain appears in Tax / Corporate \& Financial Regulation ($+61.5$\,pp) and Technology \& Privacy ($+44.4$\,pp); the smallest appears in Criminal, Drug \& State Law ($+12.5$\,pp), where many items concern well-publicized events likely covered by pretraining news. This pattern shows that the dataset's retrieval-dependence is concentrated in fast-moving regulatory domains.

\paragraph{Retrieval-depth ablation (Bar Exam).} On a 50-question subsample of Bar Exam QA, single-query web retrieval at varying depths gave: $k = 1$ snippet $\to$ 66.0\% accuracy, $k = 5$ and $k = 10$ both 62.0\%. Additional snippets add noise on reasoning-focused questions. A basic vs.\ enhanced deep-search comparison at $k = 5$ produced 64.0\% for both, but deep search incurred $40\times$ higher latency (30.4\,s vs.\ 0.75\,s).

\end{document}